\documentclass{article} 
\usepackage{aaai25}  
\usepackage{times}  
\usepackage{helvet}  
\usepackage{courier}  
\usepackage[hyphens]{url}  
\usepackage{graphicx} 
\usepackage{natbib}  
\usepackage{caption} 
\usepackage{algorithm}
\usepackage{algorithmic}
\usepackage{amsmath}
\usepackage{newfloat}
\usepackage{listings}
\DeclareCaptionStyle{ruled}{labelfont=normalfont,labelsep=colon,strut=off} 
\floatstyle{ruled}
\newfloat{listing}{tb}{lst}{}
\floatname{listing}{Listing}
\newcommand{\cbar}{\, | \,} 

\newcommand{\PROB}[1]{\mbox{Prob}\left\{#1 \right\} }

\newcommand{\NIL}[1]{}

\newcommand{\y}{\mathbf y}

\newcommand{\eyh}{\mathbf{\hat{y}}}
\newcommand{\tyh}{\hat{y}}

\newcommand{\z}{\mathbf z}

\newcommand{\e}{{\mathcal E}}   

\newcommand{\Nor} {\mathcal{N}}

\newcommand{\s}{\mathbf s}

\title{Ethics2vec: aligning automatic agents and human preferences}
\author{
    Gianluca Bontempi
}
\affiliations{
     Machine Learning Group\\
    Université Libre de Bruxelles, Belgium\\
    mlg.ulb.ac.be

}

\usepackage{bibentry}

\begin{document}

\maketitle
\begin{abstract}
%
%
The interaction of humans and intelligent agents continues to grow and will be inevitable in the near future.
Though intelligent agents are supposed to improve human experience  (or make it more efficient) it is hard from a human perspective to grasp the ethical values which are explicitly or implicitly embedded in an agent behaviour.
This is the well-known problem of \emph{alignment} which refers to the challenge of designing  AI systems which align with human values, goals and preferences.
This problem is particularly challenging since most human ethical considerations refer to \emph{incommensurable} (i.e. non-measurable and/or incomparable) values and criteria. Consider, for instance, a medical agent prescribing a treatment to a cancerous patient. 
How could it take into account (and/or weigh) incommensurable aspects like the value of a human life and the cost of the treatment?
 
Now, the alignment between human and artificial values is possible only if we define a common space where a metric can be defined and used. This paper proposes to extend to ethics the conventional Anything2vec approach, which has been successful in plenty of similar and hard-to-quantify domains (ranging from natural language processing to recommendation systems and graph analysis). 

This paper proposes a way to map an automatic agent decision-making (or control law) strategy to a multivariate vector representation, which can be used to compare and assess the alignment with human values. The rationale is that if an automatic agent implements a decision-making strategy, this strategy is optimal with respect to some loss function. At the same time, if the human accepts to adhere to the agent strategy,  this implicitly means that such agent strategy is also optimal wrt to a weighted sum of human criteria. By making such an assumption, it is possible to recover some constraints on the weights of the human criteria that the adoption of the agent strategy implies. 

The Ethics2Vec method is first introduced in the case of an automatic agent performing binary decision-making. Then, a vectorisation of an automatic control law (like in the case of a self-driving car)  is discussed to show how the approach can be extended to automatic control settings.

\end{abstract}

\section{Introduction}

Humans will be more and more exposed to interacting with automatic agents.
While this is supposed to improve the effectiveness and performance of some human tasks, it raises serious issues, mainly related to the lack of observability of the ethical values of those agents.
This is particularly sensible in decision settings (e.g. safety critical) where humans perceive incommensurability between the goal of automatic agentship (e.g. improve performance) and the related risks (e.g. impact on human safety, animal welfare or environmental quality).

In ethics, the concept of incommensurability is well known~\cite{andersson2022value} and refers to the idea that certain values or goods cannot be compared or measured objectively. This means that we cannot establish a common scale to evaluate different options or values. For example, how can we evaluate an action's consequences that affect both health and economic well-being?

Incommensurability is related to the concept of \emph{ethical pluralism}\footnote{\url{https://plato.stanford.edu/entries/value-pluralism/}}, which suggests that multiple criteria are important and cannot be reduced to a single scale of value. This means that we often have to make difficult choices between different values that are important to us.
A well-known example in automatic classification (one of the basic capabilities of an automatic agent) is the lack of commensurability of the cost of a false positive and a false negative.

There are different approaches to addressing incommensurability in ethics, including \emph{consequentialism}\footnote{\url{https://plato.stanford.edu/entries/consequentialism/}}, which seeks to evaluate the consequences of actions and choose the option that maximises overall well-being, and
\emph{deontology}, which focuses on moral rules and duties, and seeks to determine what is right or wrong based on moral principles.
If old fashioned AI was aiming at finding rules to align automatic agents (like robot) with human sensibility, modern AI and notably ML seem to be much more in line with a consequentialist approach, where the behavior of the agent emerges from the optimisation of some cost function (e.g. the classification error in a binary decision maker).

A more quantitative field of study, which is also deeply concerned with incommensurability, is multi-criteria decision making (MCDM)~\cite{triantaphyllou2000multi} that deals with making decisions when there are multiple criteria or objectives to consider. In MCDM, incommensurability refers to the challenge of comparing and evaluating different criteria that are measured in different units or have different scales.
The multiple criteria can be quantitative or qualitative, and they may have different levels of importance. MCDM is used in a wide range of fields, including business, engineering, healthcare, and environmental ~\cite{sahoo2023comprehensive}.
For example, in evaluating different transportation options, the criteria might include cost (measured in dollars), travel time (measured in minutes), and environmental impact (measured in tons of CO2 emissions). These criteria are incommensurable because they are measured in different units and cannot be directly compared.
There are several approaches to addressing incommensurability in MCDM, including 
\begin{itemize}
    \item  Weighted sum approach: it involves assigning weights to each criterion based on its relative importance, and then calculating a weighted sum of the scores for each option.
    \item Multi-attribute utility theory: it involves defining a utility function for each criterion, and then aggregating the utilities across criteria to obtain an overall utility score for each option.
\item Outranking methods: they involve comparing options pairwise and determining which option is preferred based on a set of criteria.
\end{itemize}

While several approaches have been proposed in MCDM to deal with incommensurability, such approaches concern the design phase of an automatic decision maker and not the reverse-engineering of its latent ethical values from observations.
This is indeed the aim of this paper, which takes an Anything2Vec approach to quantify the degree of human/agent alignment~\cite{christian2021alignment}.
Anything2Vec~\cite{amara2021network} is a family of machine learning approaches (e.g. Word2Vec, Node2Vec, Entity2Vec) that aim to represent complex data structures, such as words, nodes, or entities, as vectors in some space. 
Anything2Vec approaches have many applications, including natural language processing, recommendation systems and graph analysis.
These approaches are based on the idea that similar objects should be mapped to nearby points in the vector space.
Such an approach has several key benefits, like capturing complex relationships from observational data and dimensionality reduction.
Both aspects are important in the ethical domain, where it is hard for humans to assess and compare automatic agents on the basis of their ethical profile.

This paper combines the consequentialist interpretation of ethics and the MCDM weighted sum approach to define a mapping from the ethical space of an automatic black-box agent to that of the human user.
The rationale is that if the agent adopts a certain decision-making strategy, this means that such a strategy is optimal wrt the weighted sum of some agent criteria. Note that the agent criteria are not necessarily the ones of the human user. However, if the human endorses the agent decisions,  this implicitly means that the agent strategy is also optimal with respect to the weighted sum of human criteria. By making the assumption, it is possible to derive some constraints on the weights of the human criteria that the endorsement of the agent strategy implies. 

Let us illustrate this with a practical case. A self-driving car has been trained by riding millions of kilometres in the deserts of Arizona (US), where the risk of  collision is quite low. Now, suppose that the same self-driving car is used in downtown Naples (Italy), where the nature of traffic is quite different and human density is much higher.
If the Italian driver accepts the driving style of the Arizona-trained car, she is implicitly putting a lower weight on human life with respect to the original intentions of the US car designers.

The paper first considers a binary decision-making agent and shows how it is possible to reconstruct a two-dimensional vector representing the relative losses of false-positive and false-negative decisions from the agent decisions. The approach is then extended to a continuous case.
Some simulated examples illustrate the interest of the approach in simple settings.

\section{Reconstructing the ethics of a binary agent}
Let us  consider an automatic agent who is asked to perform a binary action $\hat{y}$ out of two alternatives $\{0,1\}$ 
on the basis of a given set of observations $x$ (e.g. sensor readings) associated with an unknown state of the world. 
This setting is simple but often representative of many agentic applications where the unknown state is binary (pedestrian/bag, spam/ham, fraud/genuine, health/sick, safe/unsafe), and a binary action (brake/keep going, remove/keep, block/process, prescribe a treatment or not, raise an alert or not) is required from the agent.

Suppose that the observations $x$ provide only partial information about the latent $y$. This implies that for a given $x$ the correct action $y$ a stochastic function of $x$
 described by a conditional probability.
If the agent does not perform the appropriate action $\hat{y}=y$, some cost is incurred, which is described by the following loss matrix

\begin{center}
\begin{tabular}{c||c|c}
Action & State $y=0$ & State $y=1$\\
\hline
$\hat{y}=0$ &  $L_{TN}=0$    & $L_{FN}$\\
$\hat{y}=1$ & $L_{FP}$ & $L_{TP}=0$\\
\end{tabular}
\end{center}

where $L_{FN}>0$ ($L_{FP}>0$) denotes the loss incurred if we take the action $\hat{y}=0$ ($1$) when the state is $y=1$ ($y=0$)\footnote{We assume that $y=0$ (y=1)$ $ correspond to the negative (positive) label}.
Those losses are the famous \emph{False Negative} and \emph{False Positive} losses encountered in any binary classification task.
To simplify the loss distribution, we suppose that for a given state configuration, there exists the "right action," i.e., the action that returns a null cost (e.g., we have no loss if an email is spam and we delete it).

If the agent is a black box, human users have no access to the loss matrix used for designing it. The loss matrix embodies the ethics of the agent (or its designer) but human users has no way to check or assess whether this ethics is aligned with their own values.
This paper discusses how it is possible to get some insight into the terms of the loss matrix by using the derivatives of the error rates of the agent.


 %
 Suppose that the automatic agent chooses the action to perform based on a score $\s$ for a given input $x$ as follows:
 \begin{equation}
 \label{eq:strategy}
\tyh(x)=
 \begin{cases}
 1, \quad \mbox{ if } s(x) \ge \tau^* \\
 0, \quad \mbox{ else }
 \end{cases}
 \end{equation} 

The agent threshold $\tau$ determines the agent action and the agent average loss 
\begin{multline}
\label{eq:ROCcost0}
L(\tau)=L_{FP} \cdot \PROB{\eyh=1, \y=0 \cbar \tau} +  \\+L_{FN} \cdot \PROB{\eyh=0 ,  \y=1 \cbar \tau } =
\\=L_{FP} \cdot \PROB{\eyh=1 \cbar \y=0 , \tau} \PROB{\y=0} 
+\\+  L_{FN} \cdot \PROB{\eyh=0 \cbar \y=1 ,  \tau } \PROB{\y=1} 
\end{multline}
The average loss can be approximated by using the False Positive Rate 
$$ \text{FPR} \approx  \PROB{\eyh=1 \cbar \y=0}$$
and the False Negative Rate,
$$ \text{FNR} \approx  \PROB{\eyh=0 \cbar \y=1}$$ 
well-known quantities in binary classification since they are used to trace the ROC curve~\cite{Cali15}. The ROC curve is the  set of points
 \begin{equation}
 \label{eq:nonparROC}
 \text{ROC}(\cdot)=\{ \text{FPR}(\tau), \text{TPR}(\tau), \quad \tau \in (-\infty,+\infty) \}
 \end{equation}
 where $ \text{FPR}(\tau)=\PROB{\s \ge \tau \cbar \y=0}$ and $\text{TPR}(\tau)=1- \text{FNR}(\tau)= \PROB{\s \ge \tau \cbar \y=1}$

From~\eqref{eq:ROCcost0} we write
\begin{multline}
\label{eq:ROCcost}
L(\tau)
\approx L_{FP} \cdot \text{FPR}(\tau) \cdot \hat{P}_N + L_{FN} \cdot \text{FNR}(\tau)  \cdot \hat{P}_P=\\=
L_{FP} \cdot \text{FPR}(\tau)  \cdot \hat{P}_N + L_{FN} \cdot (1-\text{TPR}(\tau) ) \cdot \hat{P}_P
\end{multline}
where $\hat{P}_N \approx \PROB{\y=0} $ ($\hat{P}_P \approx \PROB{\y=1} $) is a frequency-based estimate of the a priori probability of the negative (positive) class. Such quantities are easy to estimate and represent the (un)balancedness of the discrimination task.

If the agent strategy~\eqref{eq:strategy} is associated with the threshold $\tau^*$, this means that from the agent's ethical perspective, this threshold corresponds to an optimum, i.e. to the threshold 
\begin{equation}
\label{eq:taustar}
\tau^*=\arg \min_{\tau} L(\tau) 
\end{equation}
minimising the average cost~\eqref{eq:ROCcost}.

The agent threshold $\tau^*$  satisfies then
 $$\frac{dL}{d \tau} \Big |_{\tau^*}=L_{FP}  \frac{d \, \text{FPR}}{d \tau} \hat{P}_N - L_{FN} \cdot \frac{d \, \text{TPR}}{d\tau} \hat{P}_P=0$$
 and consequently
\begin{equation}
\label{eq:besttau}
 \frac{ d \, \text{TPR}}{  d \, \text{FPR}}\Big |_{\tau^*}=\frac{L_{FP}}{L_{FN}} \frac{\hat{P}_N}{\hat{P}_P}
 \end{equation}
 Note that the left-hand term denotes the derivative of the ROC curve~\eqref{eq:nonparROC} at the operating point associated with the threshold $\tau^*$ implemented by the agent decision strategy.  So, once the quantity on the left is estimated, we estimate the ratio of the losses of a False Positive and a False Negative. 

\subsection{Estimating the derivative}
The quantity~\eqref{eq:besttau} may be estimated in a nonparametric or parametric manner. The nonparametric approach relies on first tracing the empirical ROC curve by plotting a set of points $\langle \text{TPR}(\tau), \text{FPR}(\tau),\rangle$ for a set of values of $\tau$. Once the set of pairs is computed, a numerical differentiation technique returns an estimate of~\eqref{eq:besttau}.

The most common parametric approach~\cite{colak2012comparison} is based on the \emph{binormal} hypothesis.
 According to this hypothesis the conditional distribution
 of the score $\s$ for $\y=1$ is Normal with mean $\mu_1$ and variance $\sigma^2_1$ and the conditional distribution
 of the score $\s$ for $\y=0 $ is Normal with mean $\mu_0$ and variance $\sigma^2_0$.
 In other words, 
 \begin{align}
 \label{eq:ROCpar}
 &\PROB{\s < \tau \cbar \y=1} =\Phi\left( \frac{\tau-\mu_1}{\sigma_1}\right), \\
 &\PROB{\s < \tau \cbar \y=0} =\Phi\left( \frac{\tau-\mu_0}{\sigma_0} \right)
 \end{align}
 where the $\Phi$ function is a standard normal function.
From~\eqref{eq:ROCpar} it follows
\begin{multline}
\label{eq:ROCparTPR}
\text{TPR}=\PROB{\s > \tau \cbar \y=1}=\\=1-\PROB{\s \le \tau \cbar \y=1}=\\
=1-\PROB{\frac{\s-\mu_1}{\sigma_1} \le \frac{\tau-\mu_1}{\sigma_1}}=\\=
1-\Phi\left( \frac{\tau-\mu_1}{\sigma_1} \right) = \Phi\left( \frac{\mu_1-\tau}{\sigma_1} \right)
\end{multline}
and

\begin{multline}
\label{eq:ROCparFPR}
\text{FPR}=\PROB{\s > \tau \cbar \y=0}=1-\PROB{\s \le \tau \cbar \y=0}=\\
=1-\PROB{\frac{\s-\mu_0}{\sigma_0} \le \frac{\tau-\mu_0}{\sigma_0}}=\\=
1-\Phi\left( \frac{\tau-\mu_0}{\sigma_0} \right) = \Phi\left( \frac{\mu_0-\tau}{\sigma_0} \right)
\end{multline}


Since  
$$\frac{ d \, \text{TPR}}{  d \, \text{FPR}}=\frac{\frac{d \, \text{TPR}}{d \tau}}{ \frac{d \, \text{FPR}}{d \tau}}
$$
from~\eqref{eq:ROCparTPR} and~\eqref{eq:ROCparFPR} we obtain
\begin{equation}
\label{eq:besttauP}
\frac{ d \, \text{TPR}}{  d \, \text{FPR}} \Big|_{\tau}=
 \frac{p_{\z} \left( \frac{\mu_1-\tau}{\sigma_1} \right) }{p_{\z} \left( \frac{\mu_0-\tau}{\sigma_0} \right)} 
 \end{equation}
 where $\z \sim \Nor(0,1)$ is the standard normal variable.
By estimating $\mu_1, \sigma_1, \mu_0, \sigma_0$ from a set of observations it is possible to estimate~\eqref{eq:besttau} with~\eqref{eq:besttauP}.

\subsection{Ethics2Vec in a binary agent}

From~\eqref{eq:besttau}, the ratio between the agent False Positive and False Negative losses can  be written as 
\begin{equation}
\label{eq:besttau2}
\frac{L_{FP}}{L_{FN}}= \frac{ \frac{d \, \text{TPR}}{d \tau} \Big |_{\tau^*} }{  \frac{d \, \text{FPR}}{d \tau} \Big |_{\tau^*}}  \frac{\hat{P}_P}{\hat{P}_N} 
\end{equation}
where the numerator (denominator) represents the rate of the change of the True Positive Rate (False Positive Rate) at $\tau=\tau^*$.

This formula provides an operational manner to reconstruct the unobservable agent ethics from observational quantities, notably the rate of change of the TPR and FPR quantities.
The vector 
\begin{equation}
\label{eq:e2vB}
E=\left[\frac{d \, \text{TPR}}{d \tau} \Big |_{\tau^*},  \frac{d \, \text{FPR}}{d \tau} \Big |_{\tau^*} \right] 
\end{equation}
is then the vector which can be used to transform the unknown ethical behaviour of an agent into a quantifiable vector.

In practice, this vector could be used to measure how the automatic agent quantifies the losses associated to events which are often considered incommensurable from a human perspective. Think, for instance, of a self-driving car which has to decide whether to brake ($\eyh=1$) or not ($\eyh=0$) on the basis of an uncertain sensorial input $x$. 
The ratio~\eqref{eq:besttau2} quantifies how much the agent values the human life with respect to a non efficient (or uncomfortable) action, like braking for a worthless object. 

\subsection{Experiments with a binary agent}
This section presents a simulated experiment to show how to reconstruct the ratio~\eqref{eq:besttau}  in an observational setting.

We consider a binary classification task, and we suppose that the agent designer defines the quantities $L_{FP}$ and $L_{FN}$. 
The agent design procedure defines the optimal threshold 
$\tau^*$ of the agent by minimising the average loss (Equation~\eqref{eq:taustar}): this minimisation is possible thanks to the definition of $L_{FP}$ and $L_{FN}$.
Note that those quantities are known (implicitly or explicitly) to the agent designer but unknown to the final user who is indeed interested in knowing how the agent is aligned to her values, e.g. the loss due to a false positive wrt to that of a false negative.

Suppose that the user interacts with the agent a certain number of times
and creates a test set containing the agent's input, the action performed by the agent, and the correct action. By using such a test
set, she can compute either the parametric or nonparametric ROC curve and proceed to the estimation of the term in~\eqref{eq:besttau}. 

We consider 20 agents whose optimal thresholds have been chosen on the basis  20 different $L_{FP}/L_{FN}$ ratios\footnote{The  ratios smaller than one are $\{0.10, 0.14, 0.19, 0.23,
0.28, 0.32, 0.37, 0.41, 0.46, 0.50\}$ and the ones larger than one are  $\{2.00, 2.33 , 2.67 ,3.00,
3.33, 3.67, 4.00, 4.33, 4.67, 5.00\}$}.

\begin{figure}
\centering
    \includegraphics[width=0.35\textwidth]{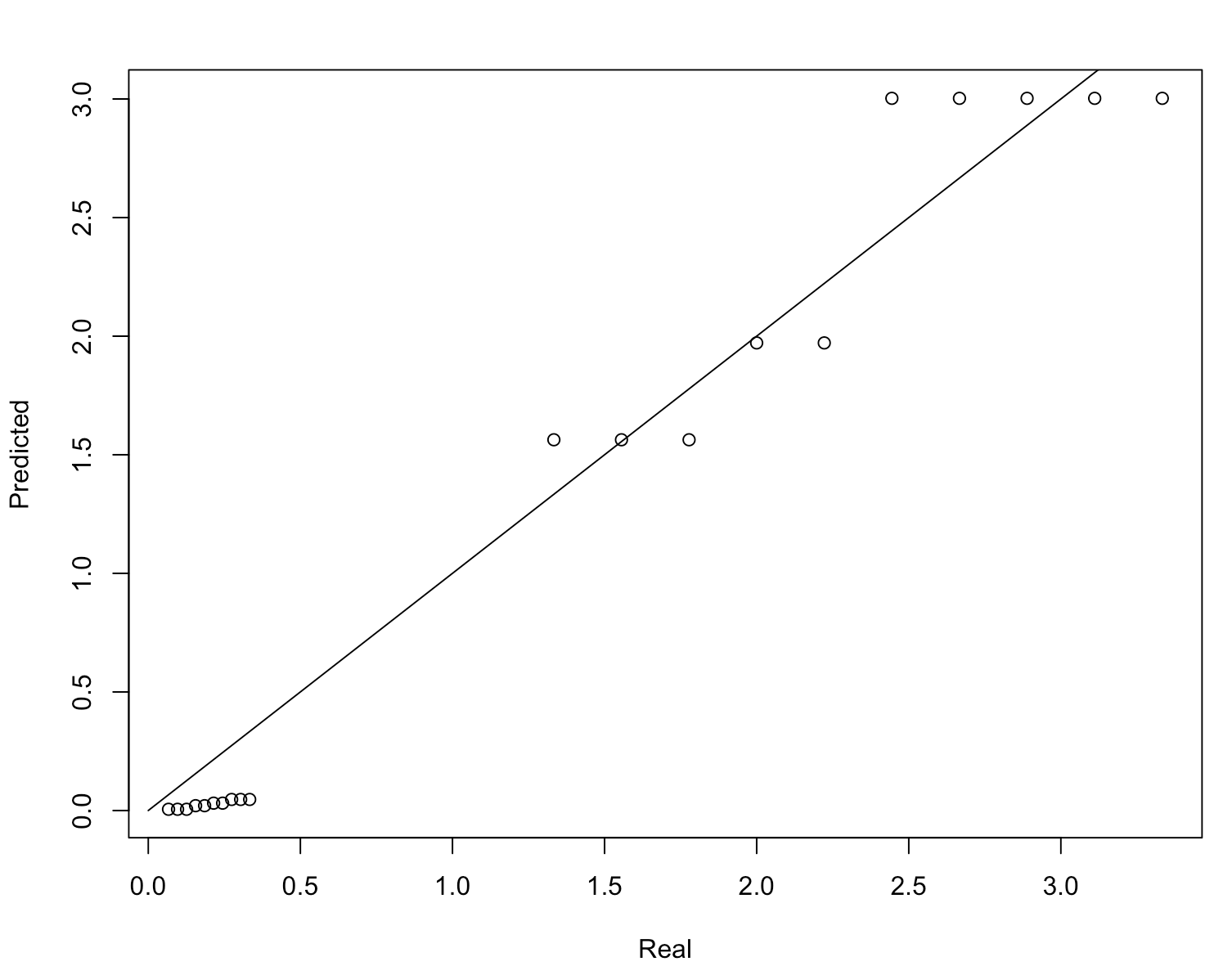}
    \caption{Real vs predicted ratios using the parametric method~\label{fig:predratio}}
\end{figure}

Figure~\ref{fig:predratio}  shows that it is possible to reconstruct the quantity~\eqref{eq:besttau} by using the parametric approach on the test set.

\begin{figure}
\centering
    \includegraphics[width=0.35\textwidth]{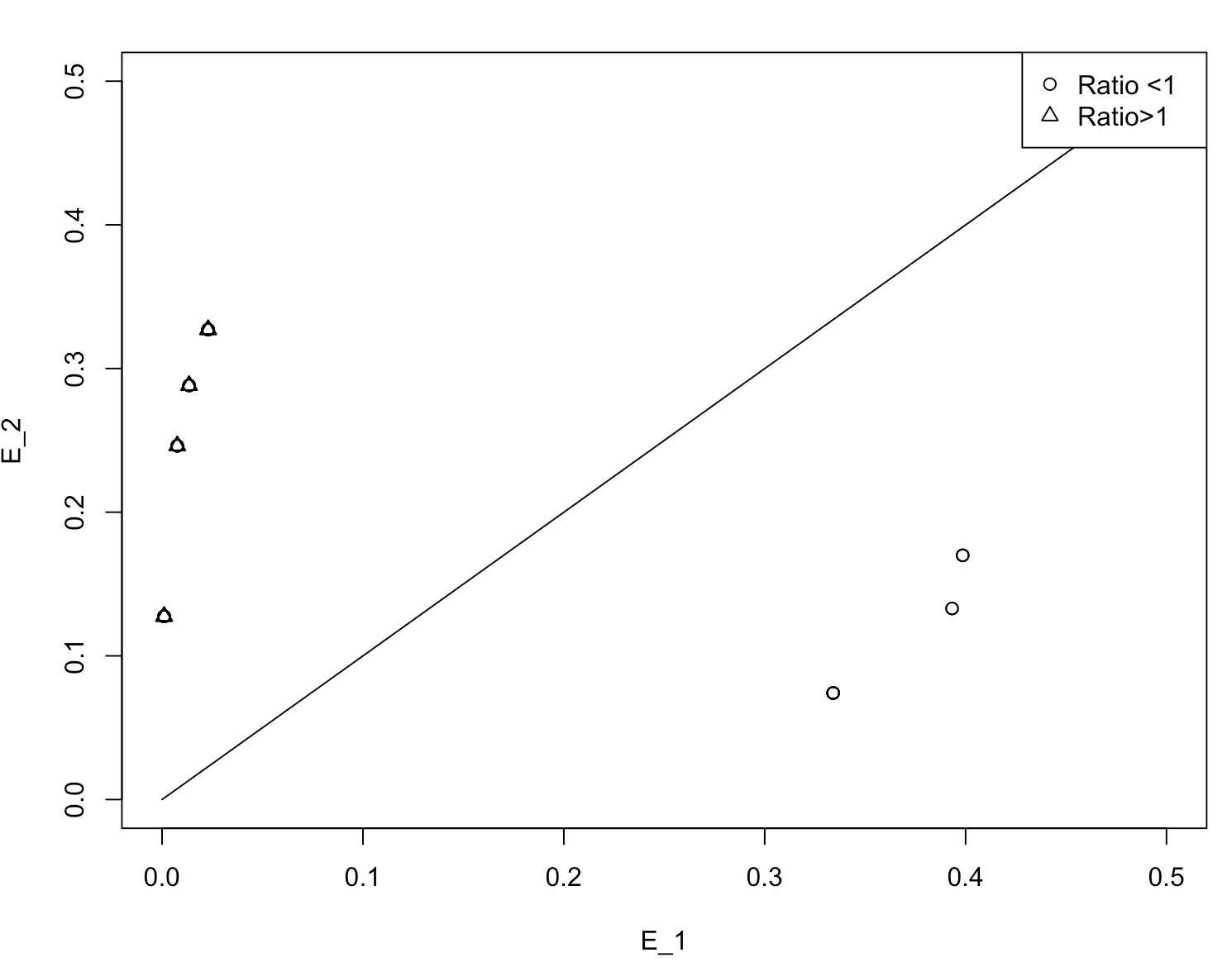}
    \caption{Ethics2vec representation of the estimated ethical weights for agents with different loss ratios~\label{fig:E2V}}
\end{figure}

Figure~\ref{fig:E2V} use the \eqref{eq:e2vB} representation of the agents' ethics. It is interesting to see how agents giving a lower loss to False Positives appear on the top-left part of the diagram, while agents giving a higher loss to False Positives are on the bottom-right part of the diagram.

\section{Reconstructing the ethics of a continuous agent}
The previous section considered a binary agent who is supposed to return, for each sensory input $x$, one action out of two options (e.g. brake or keep going).

Let us consider now an automatic agent choosing the action to perform within a continuous set on the basis of a specific control law. 
Suppose we are in a discrete time setting and that the agent performs at time $t$ a control action 
\begin{equation}
\label{eq:law}
    u(t)=K(x(t))
\end{equation}
on the basis of some observation $x(t)$. Let us focus on a finite horizon $t=1,\dots,T$.

The control law of the agent can be the result of some expert-based know-how (e.g. a rule-based controller), a conventional control law (e.g. a PID controller) or the outcome of a learning process (e.g. a neural network controller). Whatever the origin of the control law, this will determine the behaviour and implement the ethics of the automatic agent.

How a human user can assess whether the resulting behaviour is compliant with her expectations and ethical values? In the binary case, we showed that the derivative of the probability of a False Positive and of a False Negative with respect to the threshold may help quantify the ethical stance of the agent.  In this section, we propose an extension of this binary Ethics2Vec approach to the continuous case by quantifying the ethics of a controller in terms of the derivatives of a number of risk terms.

Consider, for instance, a self-driving car implementing a speed control law which determines the actual speed of the car.
Since this control law has been determined as a result of a design process, we can assume that such law is somewhat optimal with respect to some criteria. Are those the same criteria which are relevant for the final human user? is it possible to know the weights of those criteria in the global cost function which has been minimised?

We propose a strategy to map the control law of the agent to a vector of values which are relevant for the final user. If in the case of classification, the two relevant coordinates are the risk of false positives and false negatives, respectively, here we define a set of risk variables which are relevant from a human perspective.
Examples of risk variables in the case of a self-driving car could be the risk of being late, the risk of an uncomfortable experience for the passengers, the risk of an accident or the risk of being fined by the traffic police.

Let $R$ be the number of considered risks, 
where
$$r_i(x(t), u(t))=\PROB{\e_i \cbar u(t),x(t)} \quad i=1,\dots,R$$ denote the 
conditional probability\footnote{For the sake of simplicity, we assume here that the risk is stationary, i.e. it is not explicitly dependent on $t$.}that the event $\e_i$ take place at time $t$ given the current observation $x(t)$ of the state and the current control action $u(t)$. 

For instance the $\e_i$ could represent an accident and $r_i(t)$ the probability that an accident takes place at time $t$ given the current speed $u(t)$ and the congestion state of the road represented by the vector $x(t)$: for instance, in this case $x(t)$ could be the image perceived by the LIDAR sensor of the car.

Analogously to~\eqref{eq:e2vB}, we define 
\begin{equation}
\label{eq:e2vC}
E(t)=\left[\frac{d r_1(x(t),u)}{d u}\Big|_{u=K(x(t))},\dots, \frac{ d r_R(x(t), u)}{du }\Big|_{u=K(x(t))} \right]
\end{equation}
the Ethics2Vec representation of the ethical behaviour at time $t$.

The justification of this approach is that, if the user adopts the control law~\eqref{eq:law}  implemented in the agent, according to a consequentialist ethical approach, this law should optimise some loss expressed as a function (e.g. weighted sum) of human ethical criteria.

If we adopt an MCDM weighted sum approach and define the  loss for the user as the weighted sum 
$$ L= \sum_{t=1}^T \sum_{i=1}^R w_i r_i ( x(t), u(t))$$
of the  set $\{ r_i \}$ of risks, the agent control action must satisfy the condition
\begin{multline}
\frac{d L } {du }=0  \Rightarrow \sum_{t=1}^T \sum_{i=1}^R w_i \frac{d r_i ( x(t), u)}{d u}\Big |_{u=K(x(t))}=0  \Rightarrow \\ \sum_{t=1}^T W \cdot E(t)=0
\end{multline}
where $E(t)$ is defined in~\eqref{eq:e2vC}.

Since the weights represent the relative importance of the ethical criteria in the global loss function,  the vector~\eqref{eq:e2vC} defines a set of constraints on the relevance of each criterion. For instance if $R=2$ by measuring the vector $E(t)$ for $t=1,\dots,T$ we obtain that the weights $w_1$ ad $w_2$ must satisfy the relationship
$$ \sum_{t=1}^T w_1 \frac{d r_1 ( x(t), u(t))}{d u} + w_2 \frac{d r_2 ( x(t), u(t))}{d u}=0 $$ or equivalently
\begin{equation}
\label{eq:W2}
   \frac{w_1}{w_2}= -\sum_{t=1}^{T} \frac{\frac{d r_2 ( x(t), u}{d u}\Big |_{u=K(x(t))}}{\frac{d r_1 ( x(t), u}{d u}\Big |_{u=K(x(t))}} 
\end{equation}
We invite the reader to compare the expression above with the relationship~\eqref{eq:besttau}.

\subsection{Experiment with a continuous action agent}

Let us consider a very simplistic simulation of a self-driving car
starting from the  position $x(0)=0$ and aiming to reach the destination at position $x=250$ (in kilometers) within a given time $T=4$ (in hours) . 

\begin{figure}
\centering
    \includegraphics[width=0.35\textwidth]{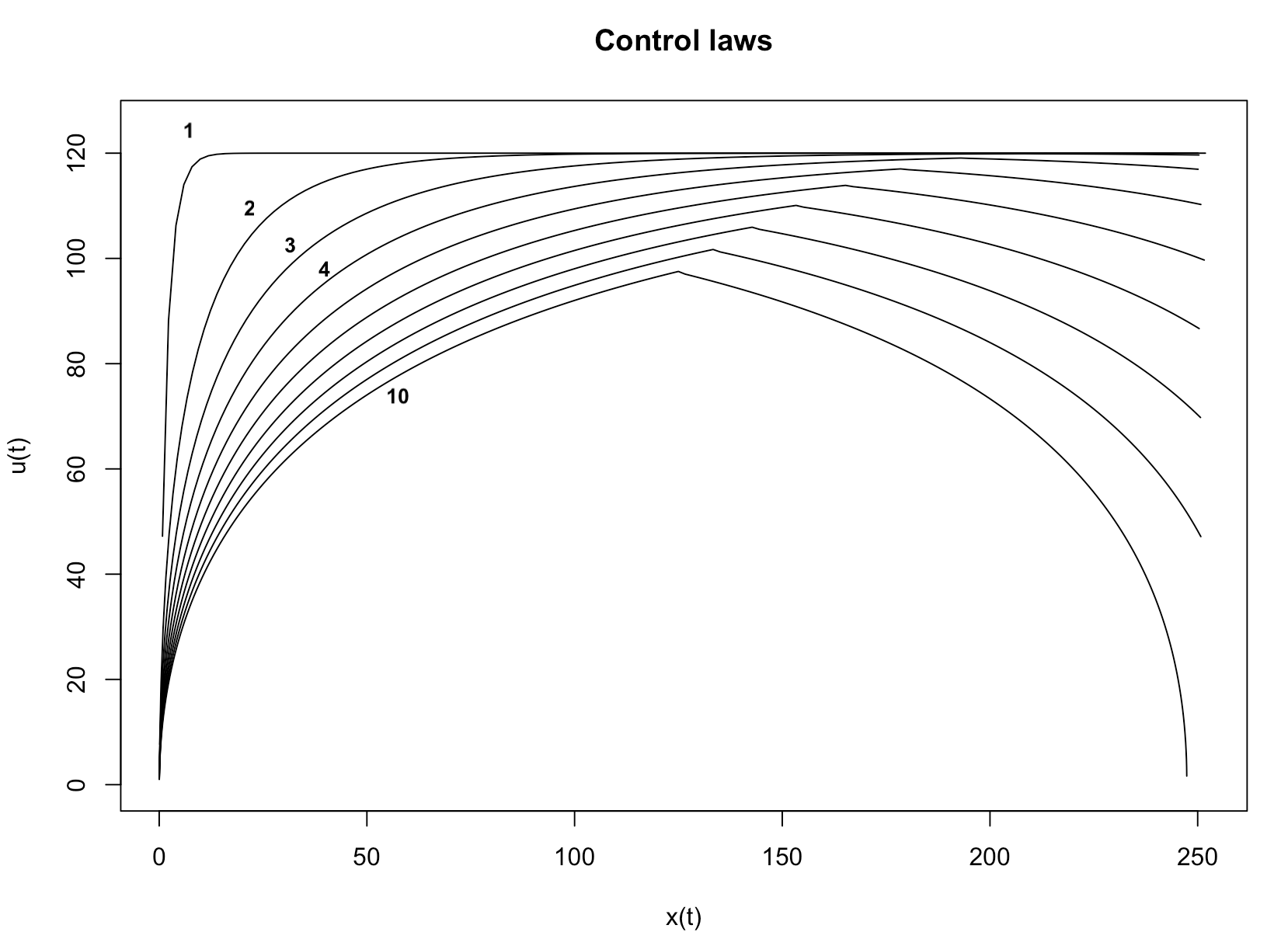}
    \caption{10 different control laws~\label{fig:claw}}
\end{figure}

Let us consider a set of 10 different control laws 
$$u(t)=K_i(x(t)), \quad i=1,\dots, 10$$
illustrated in Figure~\ref{fig:claw}
where  $x(t)$ is the current position of the car and $i$ is the index of the control law.

\begin{figure}
\centering
    \includegraphics[width=0.35\textwidth]{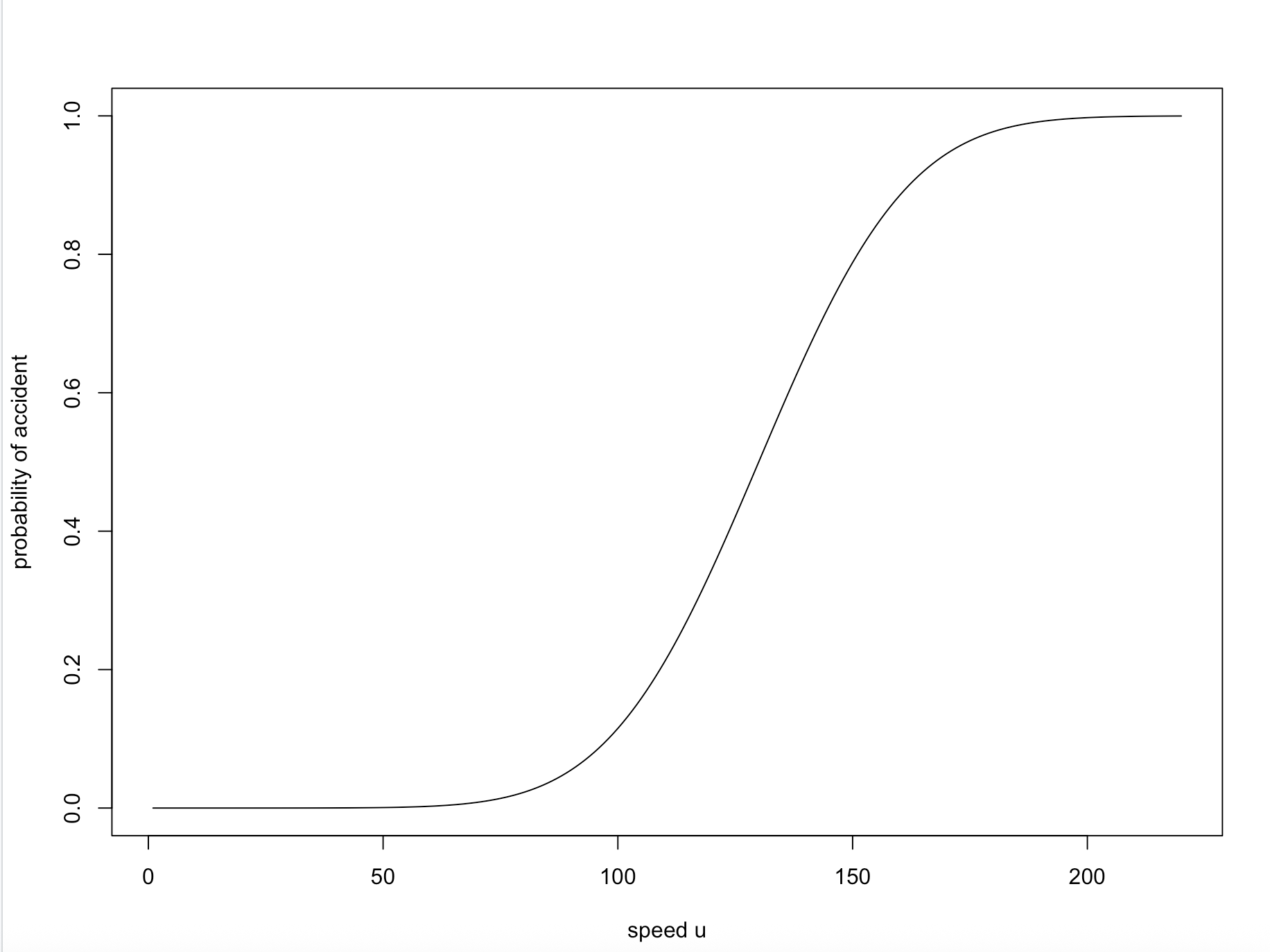}
    \caption{Probability of an accident as a function of the car speed (in km/h)~\label{fig:pacc}}
\end{figure}

\begin{figure}
\centering
    \includegraphics[width=0.35\textwidth]{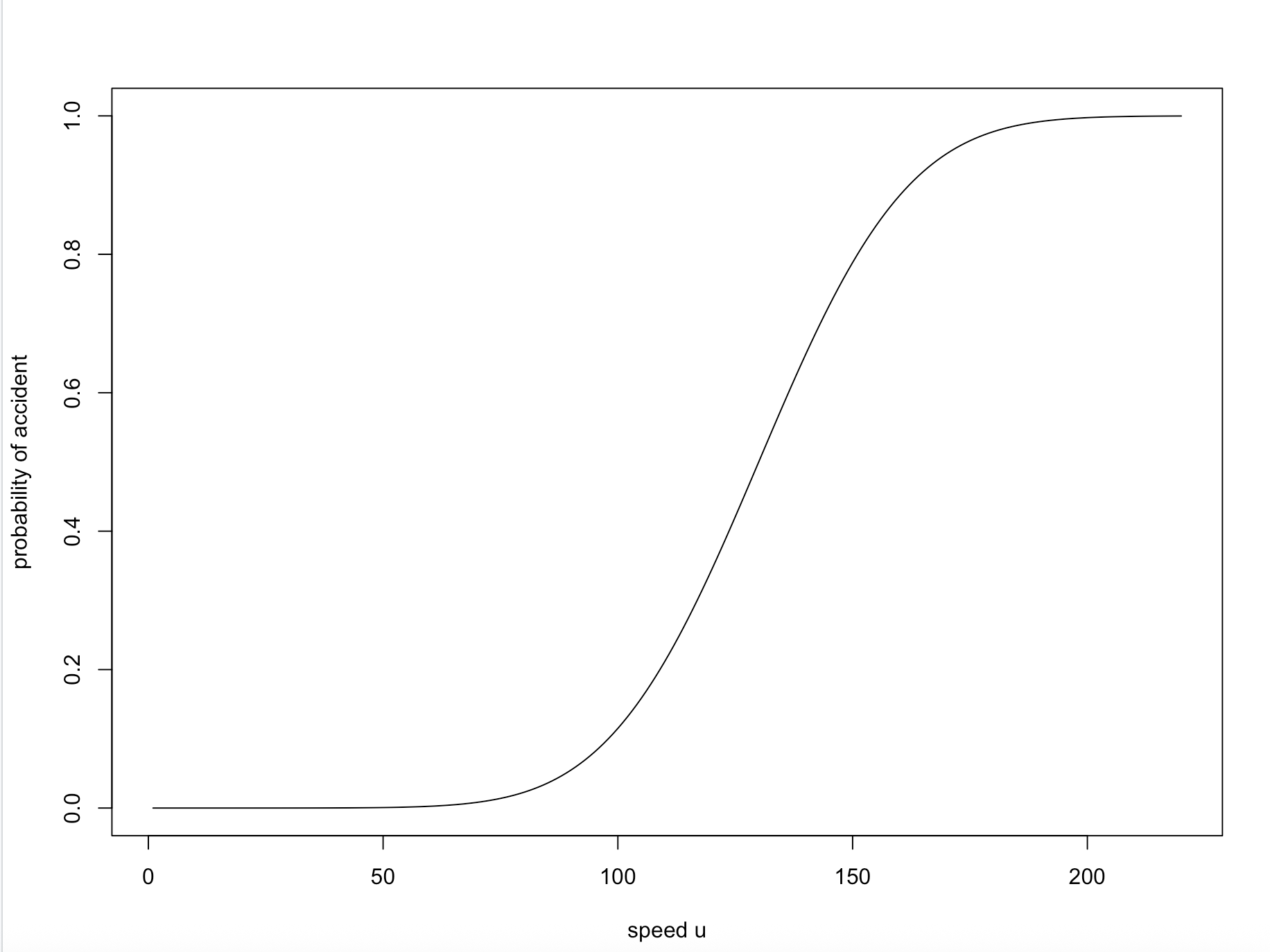}
    \caption{Probability of being late at time $t=2$ hours as a function of the estimated remaining time~\label{fig:pd}}
\end{figure}

Let $R=2$ and $r_1(u)$ represent the risk of accident (Figure~\ref{fig:pacc}) while $r_2(t)$ is the risk of being late at destination.
Note that the risk of an accident is considered as a function of the speed $u$ only, while the risk of being late depends on time $t$ and what is the estimated remaining time measured at time $t$ in position $x(t)$. Figure~\ref{fig:pd}
shows the probability of being late at time $t=2$ as a function of the estimated remaining time.

Note that, for the sake of the simulation, we suppose to we know both the probability of accidents and of being late. In a more realistic setting, those quantities could be estimated on the basis of historical data (e.g. accident road statistics) and made more complex than here (e.g. the probability of the accident could depend also on the road congestion).

\begin{figure}
\centering
    \includegraphics[width=0.35\textwidth]{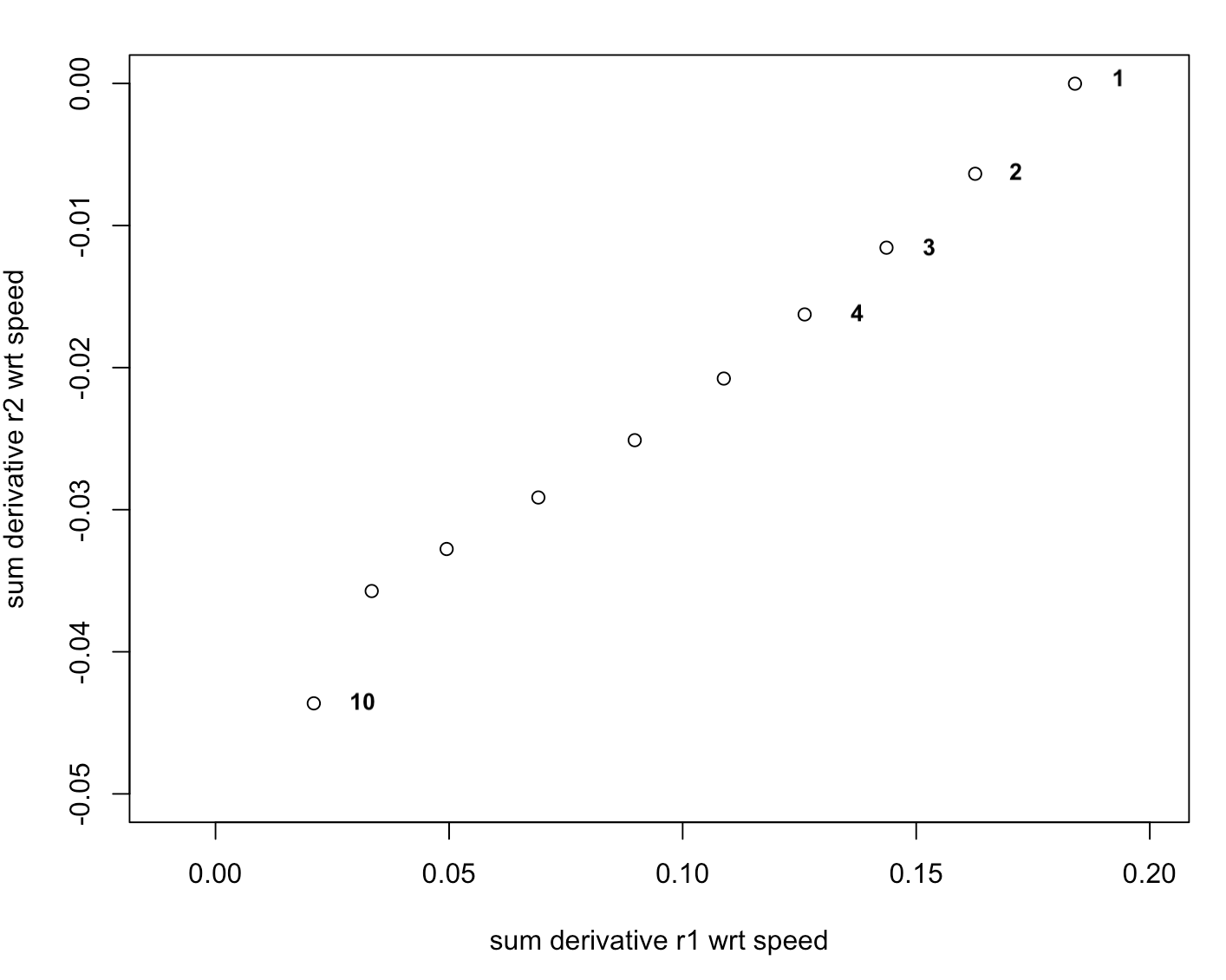}
    \caption{Ethics2vec representation of the 10 control laws~\label{fig:E2Vc}}
\end{figure}

It appears evident then that the control laws with lower indices correspond to a more aggressive
driving style, giving somewhat more importance to the fact of not being late rather than minimising the risk of an accident. The 10 control strategies are represented by two-dimensional vectors in Figure~\ref{fig:E2Vc}. The two coordinates are the two components of the vector $E$
obtained by averaging $E(t)$ (Equation~\eqref{eq:e2vC} with $R=2$) over time
($t=1,\dots,T=4)$.

From the Figure~\ref{fig:E2Vc} it appears that 
\begin{itemize}
\item the average derivatives of $r_2$ wrt speed are negative since by increasing the speed, the probability of being late decreases;
\item the average derivatives of $r_1$ wrt speed are positive since by increasing the speed, the probability of having an accident increases;
\item the most aggressive strategies (lower indices) correspond to points on the top right part of the diagram: this means that for those strategies, a large increase of the risk of the accident can be compensated even by a small reduction of the probability of being late. In other terms, being late is considered so 'ethically' bad by the driving agent that it prefers to take large risks.
\item the most careful driving strategies (highest indices) correspond to points on the bottom left part of the diagram: this means that for those strategies, a small increase in the risk accident is worth only if this induces a very large reduction of the risk of being late. In other words, the ethical consequence of causing an accident is considered so serious by the driving agent that it prefers to increase speed only when this allows an enormous gain in terms of saved time. 
\end{itemize}

We use~\eqref{eq:W2} to derive the relative weights of the two criteria for the two control laws.

\section{Conclusion}

Whatever the context, the action of an automatic agent has always some ethical relevance: for instance, a malfunctioning of a self-driving car can kill a pedestrian, and the malfunctioning of an automatic gate can kill a child\footnote{\tiny \url{https://www.nbcnews.com/video/child-killed-by-automatic-gate-in-tragic-accident-407903811753}}.
The a posterior analysis of any tragic malfunctioning of an automatic agent reminds what a computer engineer should already know, i.e. that, even for the simplest mechanism, a zero error behaviour in an uncertain setting is not possible and that the agent (implicitly or explicitly) needs to weight between costs of false positive vs costs of false negative.

But a human user (or an owner of an automatic agent) typically does not know anything about the ethical values of a black box agent. In simpler words, she does not know how the design of the agent trades some costs (e.g. bad performance or being late) with other costs (e.g. killing a person or putting the user's life at risk). This paper assumes that each automatic agent implementing a decision strategy or control law is implicitly (or explicitly) minimising a cost function where different kinds of losses are weighted and combined.  

Though the agent relevant criteria are not necessarily visible or known by the user, the user may situate the control law in its space of values and then estimate which ethical trade-off is acceptable by accepting that an automatic agent is making actions for him.
The mapping of the agent strategy in the ethical domain of the user is done by using an Ethics2Vec approach and allows a comparison of different strategies and an assessment of a possible alignment of the agent action with her own human values.

\bibliography{e2v}
\end{document}